\documentclass{article}

\usepackage[final]{neurips_2019}

\usepackage[utf8]{inputenc} % allow utf-8 input
\usepackage[T1]{fontenc}    % use 8-bit T1 fonts
\usepackage{hyperref}       % hyperlinks
\usepackage{url}            % simple URL typesetting
\usepackage{booktabs}       % professional-quality tables
\usepackage{amsfonts}       % blackboard math symbols
\usepackage{nicefrac}       % compact symbols for 1/2, etc.
\usepackage{microtype}      % microtypography
\usepackage{algorithm}      % algorithm
\usepackage{algorithmic}
\usepackage{amsmath}
\usepackage{natbib}
\usepackage[inline]{enumitem}
\usepackage[pdftex]{graphicx}
\usepackage[dvipsnames]{xcolor} 
\usepackage{siunitx}
\usepackage[normalem]{ulem}
\usepackage{amsthm}

\newtheorem{theorem}{Theorem}

\title{A study of data and label shift in the LIME framework}

\author{%
  Amir Hossein Akhavan Rahnama \\
  Department of Computer Science \\
  KTH Royal Institute of Technology \\
  Stockholm, Sweden \\
  \texttt{arahnama@kth.se} \\
  \AND
  Henrik Boström \\
  Department of Computer Science \\
  KTH Royal Institute of Technology \\
  Stockholm, Sweden \\
  \texttt{hbostrom@kth.se} \\
}
\begin{document}

\maketitle

\begin{abstract}
LIME is a popular approach for explaining a black-box prediction through an interpretable model that is trained on instances in the vicinity of the predicted instance. To generate these instances, LIME randomly selects a subset of the non-zero features of the predicted instance. After that, the perturbed instances are fed into the black-box model to obtain labels for these, which are then used for training the interpretable model. In this study, we present a systematic evaluation of the interpretable models that are output by LIME on the two use-cases that were considered in the original paper introducing the approach; text classification and object detection. The investigation shows that the perturbation and labeling phases result in both data and label shift. In addition, we study the correlation between the shift and the fidelity of the interpretable model and show that in certain cases the shift negatively correlates with the fidelity. Based on these findings, it is argued that there is a need for a new sampling approach that mitigates the shift in the LIME's framework.
\end{abstract}

\section{Introduction}
Local surrogate methods go back a long way in the literature of interpretable machine learning, see e.g. \citep{schmitz1999ann}. Local Interpretable Model-agnostic Explanations (LIME) is a recent example of these post-hoc methods, which has received significant attention \citep{ribeiro2016should}. LIME provides an explanation for a single instance using a local surrogate, where the black-box is trained on the original input space and the surrogate is trained on an interpretable space. LIME's usefulness lies in its flexibility to provide explanations for different data types, like text and images, while being model-agnostic.

In the next section, we briefly discuss some related studies. In Section 3, we present a novel approach of measuring data and label shift in LIME\footnote{LIME offers two algorithms for interpretability: Sparse Local Linear Explanations and SP-LIME. For clarity, this study considers the former only.}, and apply it to the original case studies considered in \citep{ribeiro2016should}, as described in Section 4. Finally, in Section \ref{conclusion}, we summarize the main conclusions and point out some directions for future research.

\section{Related work}
\label{related-work}

In \citep{alvarez2018robustness}, a wide range of explanation methods for image classification were investigated. The study focused on empirically investigating \textit{robustness} of explainability methods, and it showed that even minor changes in an input image can cause LIME to produce different explanations with a substantial variance in the selected (interpretable) features. The importance of locality for obtaining models with high fidelity was highlighted by \citep{laugel2018defining}. A method called \textbf{LocalSurrogate} was proposed to improve the sampling procedure of LIME. Some experiments on a set of UCI datasets were performed to show the usefulness of the approach. However, this approach still employs random perturbation, and the consequences of this choice will be investigated in this work. In \cite{zhang2019should}, the authors showed that there is a significant level of uncertainty present even in LIME's explanations of black-boxes with high test accuracy on synthetic and some UCI repository datasets. 

\section{Method}
%\label{method}

A standard assumption in machine learning is that training and test data are coming from the same distribution. One approach to detect if the assumption is violated is to calculate the divergence between the two samples and decide whether a shift has occurred, see \citep{quionero2009dataset}. The divergence between the two samples can be measured in many ways, e.g., using KL-divergence or Bregman divergence. Maximum mean discrepancy \citep{gretton2012kernel} is chosen here, since it not only allows for being computed reasonably fast, i.e., in quadratic time, but also provides an acceptance threshold that can be computed without extra constraints on the distributions being tested. In addition to that, it comes with a two sample testing framework with theoretical guarantees, see \citep{gretton2012kernel} for more details.

%\subsection{Fidelity}

Assume that we are interested in the explanation of an instance $x$ that is predicted by a black-box model $f$, for which we get a numerical score, e.g., an estimate of the class probability, with respect to some specific class label $y$, here denoted by $f_y(x)$, together with a corresponding score for the class label output by the local surrogate, here denoted $g_y(x)$. We then define the fidelity of $g$ to $f$ as follows:
\begin{equation}
\label{fidelity_regression}    
    \mathbb{F}(x,y) = \frac{1}{|f_y(x) - g_y(x)| + 1}
\end{equation}

In this work, the above formula is used for calculating the fidelity between an interpretable model and the underlying black-box model with respect to a specific set of instances. 

\section{Empirical investigation}
\label{results}
In this section, we aim to answer the following questions about the LIME framework:

\begin{itemize}
    \item Does the perturbed instances ($Z$) come from another distribution than the original training instances ($X_\text{train}$), i.e., has a data shift occurred? 
    \item If the above holds, do the black-box predictions for the two sets ($\{f(z) : z \in Z\}$ and $\{f(x) : x \in X_\text{train}\}$) come from different distributions, i.e., has a label shift occurred? 
    \item Does the above shifts have a negative effect on the fidelity of the interpretable model?
\end{itemize}

%If the answer to the first question is negative, since $f$ is trained on $X_\text{train}$, {\color{blue}{applying $f$ on LIME's perturbed samples \emph{directly}} {\color{blue}{ are from a completely different distribution}} cannot have guarantees for the accuracy of predicted labels from the blackbox model, $f$}, since it is being used for prediction of instances from another distribution. {\color{blue}{\sout{This shows that \emph{random perturbation cannot be a reliable approach to gather new samples in LIME's algorithm}}}}. If the answer to the second question is positive, {\color{blue}{then clearly there are no guarantees for the accuracy of these labels when used \emph{directly} to train an interpretable model}}.

%{\color{blue}{If the answer to the first question is negative and to the second question positive, since $f$ is trained on $X_\text{train}$, applying $f$ on LIME's perturbed samples \emph{directly}} can no longer have guarantees for the accuracy of predicted labels from the black-box model, $f$. This is of utter importance, since the interpretable model of LIME is trained on the perturbed samples $Z$, and the output of the black-box model, $f(Z)$ as labels.}

\subsection{Text classification with SVM}
\label{svm-usecase}
In this experiment, we provide explanations for all the test instances in the Newsgroups dataset with regards to the class \textit{atheism}. The aim is to show both the magnitude of divergence and frequency in which shift occurs. To investigate the local neighborhood of an explained instance $x$, $n$ neighbouring training instances (using the cosine kernel) are selected, here denoted as $X_\text{knn}$. After that, for our data shift test, we run a two sample MMD kernel test ($\alpha=0.05$) between $X_\text{knn}$ and the perturbed instances ($Z$). In this case, the null hypothesis is $H_0: P_{X_\text{knn}} = P_Z$. For the label shift test, we run a two sample MMD kernel test ($\alpha=0.05$) between $F_{X_\text{knn}} = \{f(x) : x \in X_\text{knn}\}$ and $F_Z = \{f(z) : z \in Z\}$. In this case, the null hypothesis is $H_0: P_{F_{X_\text{knn}}} = P_{F_Z}$.

In Table \ref{data-shift-table}, the frequency of accepted tests with different values of $n$ are shown. As can be seen, as the value of $n$ increases, the divergence values increases significantly. Table \ref{data-shift-table} shows that for sample sizes that are larger than 20, the first null hypothesis stated above can be rejected for \emph{all test instances}. The results of the tests for label shift are presented in Table \ref{label-shift-table}. Similar to the previous test, as the number of samples, $n$, increases, the divergence becomes larger. The increase is at a lower pace compared to the test for the data shift in this case, nonetheless with even $n=2$, more than a third of the two sample test cases can be rejected. The experiments have hence shown that the random perturbation of LIME may result in considerable data and label shift even for small sample sizes.

%\begin{figure}
%  \centering
%  \includegraphics[width=0.8\linewidth]{small_data_shift_newsgroup.png}
%  \caption{MMD divergence histogram for data shift test}
%  \label{data_shift_text_classification}
%\end{figure}

%\begin{figure}
%  \centering
%  \includegraphics[width=0.8\linewidth]{small_label_shif_test_newsgroup.png}
%  \caption{MMD divergence histogram for label shift test}
%  \label{label_shift_text_classification}
%\end{figure}

\begin{table}[t]
\caption{Data shift two sample test results for Newsgroup test instances: $H_0: P_{X_\text{knn}} = P_Z$  ($\alpha=0.05$)}
\label{data-shift-table}
\vskip 0.10in
\begin{center}
\begin{small}
\begin{sc}
\begin{tabular}{lcccr}
\toprule
$n$ & Reject & Failed to reject & MMD \\
\midrule
$2$    & 417 (57\%)& 300 (43\%) & 0.42 $\pm$ 0.34 \\
$20$    & 717 (100\%)& 0 (0\%) & 5.56 $\pm$ 1.58 \\
$100$    & 717 (100\%)& 0 (0\%) & 24.77 $\pm$ 8.00 \\
$200$    & 717 (100\%)& 0 (0\%) & 44.20 $\pm$ 15.84 \\
$500$    & 717 (100\%)& 0 (0\%) & 87.35 $\pm$ 36.75 \\
\bottomrule
\end{tabular}
\end{sc}
\end{small}
\end{center}
\vskip -0.1in
\end{table}

\begin{table}[t]
\caption{Label shift two sample test results for Newsgroup test instances along with fidelity measures:$H_0: P_{F(X_\text{knn})} = P_{F(Z)}$ ($\alpha=0.05$)}
\label{label-shift-table}
\vskip 0.10in
\begin{center}
\begin{small}
\begin{sc}
\begin{tabular}{lccccr}
\toprule
$n$ & Reject & Failed to reject & MMD \\
\midrule
$2$    & 239 (33.3\%)& 478 (66.6\%) & 0.21 $\pm$ 0.56  \\
$20$    & 515 (71.8\%)& 202 (28.1\%) & 2.38 $\pm$ 1.93 \\
$100$    & 716 (99.8\%)& 1 (0.2\%) & 11.97 $\pm$ 7.69  \\
$200$    & 717 (100\%)& 0 (0\%) & 24.29 $\pm$ 14.47 \\
$500$    & 717 (100\%)& 0 (0\%) & 63.06 $\pm$ 31.16 \\
\bottomrule
\end{tabular}
\end{sc}
\end{small}
\end{center}
\vskip -0.1in
\end{table}

%\begin{table}[t]
%\caption{Fidelity measures on Newsgroup dataset}
%\vskip 0.15in
%\begin{center}
%\begin{small}
%\begin{sc}
%\begin{tabular}{lcr}
%\toprule
%$n$ & Fidelity\\
%\midrule
%$2$    & N/A \\
%$20$    &  N/A \\
%$50$    &  0.72 $\pm$ 0.03 \\
%$100$    &  0.68 $\pm$ 036 \\
%$200$    &  0.24 $\pm$  0.031 \\
%\bottomrule
%\end{tabular}
%\end{sc}
%\end{small}
%\label{text-fidelity-table}
%\end{center}
%\vskip -0.1in
%\end{table}

In Figure \ref{fidelity_imagenet_newsgroup}, the average fidelity of the interpretable model output by LIME over all test instances is displayed for varying sample sizes. Although, there is no widely accepted fidelity criteria for surrogate models like LIME, the achieved fidelity rates can hardly be accepted as they are not much better than random.
%\footnote{function can achieve an average fidelity of 71\% over all instances in 100000 trials.}

%interpretable models (and hence features) that are learned with samples sizes more than 200 may be questioned. \footnote{It should be noted that for $n < 50$, no interpretable model is learned by \textbf{LIME}, since the small sample size is insufficient for learning in Ridge Regression and LASSO regularization path.}

%. We have identified that the interpretable model class used in LIME, namely Ridge Regression returns $W_g=0$ {\color{blue}{due to insufficient number of data points}}. The same applies to LASSO regularization path in which no active features are selected. {\color{blue}{This can introduce a challenge to users of \textbf{LIME}: while there is less shift in smaller sample sizes and it might be preferable to work with small perturbed samples, \textbf{LIME}'s interpretable model cannot be trained with small sample sizes in this case.}}

\subsection{Object detection with deep neural networks}
\label{object_detection_dnn}
 In this use-case, LIME explanations are calculated for the top-1 predicted class label of each test instance test instances in ImageNet (\citep{deng2009imagenet}). Due to the fact that it is computationally infeasible to find nearby instances via K-Nearest Neighbours directly in the ImageNet training dataset, an alternative approach is considered here: for each explained instance ($x$), a random sub-sample from the images with the class equal to the predicted class of the black-box model, namely $F(x)$ is compared against the perturbed samples \textbf{LIME}, namely $Z$. This random sub-sample is called $X_{\text{local}}$ and since $F$ is an accurate black-box model trained on $X_\text{train}$ and $x \in X_\text{train}$, therefore our assumption is that the predicted value of $F(x)$ can help to find similar instances to $x$ in $X_\text{train}$ without the need for knowing the ground truth with only negligible error. The test layouts for both data and label shift are exactly equal to those in \ref{svm-usecase}, if the notation of $X_\text{knn}$ is replaced with $X_{\text{local}}$.

\begin{figure}
  \centering
  \includegraphics[scale=0.20]{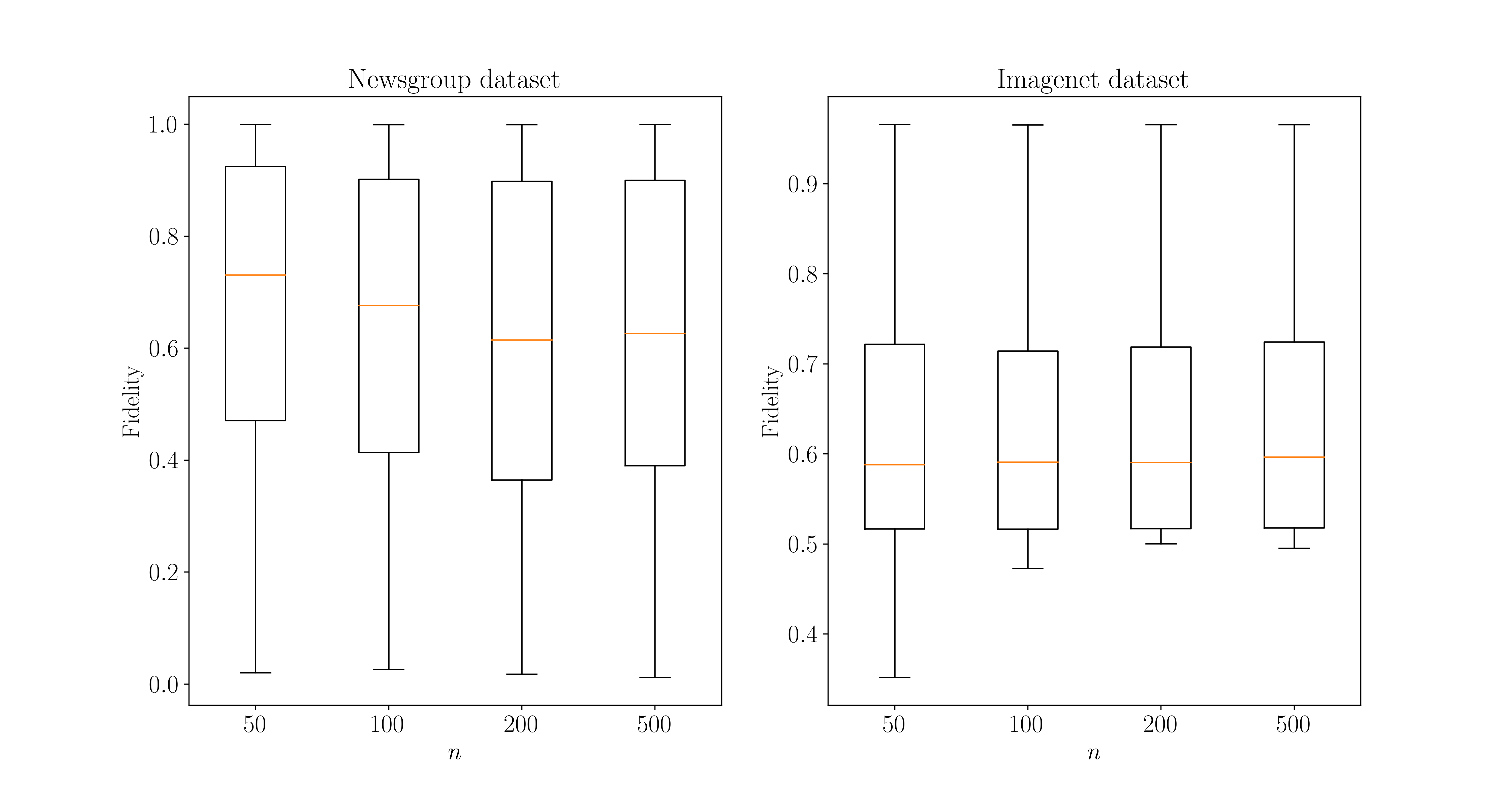}
  \caption{Fidelity and MMD divergence in Newsgroup and ImageNet dataset}
  \label{fidelity_imagenet_newsgroup}
\end{figure}

In this use-case, due to our limited computational budget, we have performed the tests on a sub-sample of 200 instances in \textbf{Imagenet}\footnote{See additional material for the predicted class distribution of this sample}. In Table \ref{data_shift_object_detection_table} and Table \ref{label_shift_object_detection_table}, the results of our experiments are shown. In both experiments, the mean divergence of data and label shift and the number of rejected two sample tests are larger when compared to the the former use-case. As before, even with small number of perturbations, significant data and label shifts are visible. In Figure \ref{fidelity_imagenet_newsgroup}, it can be seen that in ImageNet example, the fidelity of the interpretable model is worse than the Newsgroup use-case on average. 
%In addition, the fidelity of the interpretable model is not affected as the number of samples increase. We argue that one possible explanation is the fact that the fidelity of the interpretable model in our sub-sample is achieves a similar the accuracy of random chance on average\footnote{For example, a random function could achieve a fidelity of 0.75 on average in 1000000 trials. See additional material}. 

Due to the overall low fidelity in the ImageNet use-case study, one may argue that the quality of the output explanations of LIME for object detection using deep neural networks can be questioned.

%\begin{figure}
%  \centering
%  \includegraphics[width=0.8\linewidth]{NeuRIPS2019/small_data_shift_image_net.png}
%  \caption{MMD Divergence values for data shift in the Imagenet dataset}
%  \label{data_shift_object_detection}
%\end{figure}

%\begin{figure}
%  \centering
%  \includegraphics[width=0.8\linewidth]{NeuRIPS2019/small_label_shift_image_net.png}
%  \caption{MMD Divergence values for label shift in the ImageNet dataset}
%  \label{label_shift_object_detection}
%\end{figure}

\begin{table}[t]
\caption{Data shift two sample test results for ImageNet test instances for their top predicted class label: $H_0: P_{X_\text{local}} = P_{Z}$ ($\alpha=0.05$)}
\vskip 0.15in
\begin{center}
\begin{small}
\begin{sc}
\begin{tabular}{lcccr}
\toprule
$n$ & Reject & Failed to reject & MMD \\
\midrule
$50$    & 188 (100 \%)& 188 (0\%) & 6.56 $\pm$ 0.13 \\
$100$    & 188 (100\%)& 0 (0\%) & 13.16 $\pm$ 0.20 \\
$200$    & 188 (100\%)& 0 (0\%) & 26.21 $\pm$ 0.35 \\
$500$    & 188 (100\%)& 0 (0\%) & 65.32 $\pm$ 0.67 \\
\bottomrule
\end{tabular}
\end{sc}
\end{small}
\label{data_shift_object_detection_table}
\end{center}
\vskip -0.1in
\end{table}

\begin{table}[t]
\caption{Label shift two sample test results for ImageNet test instances for their top predicted class label: $H_0: P_{F(X_{\text{local}})} = P_{F(Z)}$ ($\alpha=0.05$)}
\vskip 0.15in
\begin{center}
\begin{small}
\begin{sc}
\begin{tabular}{lcccr}
\toprule
$n$ & Reject & Failed to reject & MMD \\
\midrule
$50$    & 188 (100 \%)& 0 (0\%) & 34.18 $\pm$ 6.13 \\
$100$    & 188 (100\%)& 0 (0\%) & 69.03 $\pm$ 12.73 \\
$200$    & 188 (100\%)& 0 (0\%) & 139.16 $\pm$ 25.63 \\
$500$    & 188 (100\%)& 0 (0\%) & 346.30 $\pm$ 67.99 \\
\bottomrule
\end{tabular}
\end{sc}
\end{small}
\label{label_shift_object_detection_table}
\end{center}
\vskip -0.1in
\end{table}

%\begin{table}[t]
%\begin{table}[!htbp]
%\caption{Fidelity measures on LIME explanations for top predicted class in ImageNet and Newsgroup dataset}
%\vskip 0.15in
%\begin{center}
%\begin{small}
%\begin{sc}
%\begin{tabular}{lcr}
%\toprule
%$n$ & Fidelity (ImageNet) & Fidelity (Newsgroup)\\
%\midrule
%$2$    & 0.013 $\pm$ 0.08 & \text{N/A} \\
%$5$    &  0.022 $\pm$ 0.11 & \text{N/A} \\
%$10$    &  0.025 $\pm$ 0.10 & \text{N/A} \\
%$20$    &  0.023 $\pm$ 0.11 & \text{N/A} \\
%$50$    & 0.022 $\pm$  0.12 & 0.72 $\pm$ 0.03 \\
%$100$    &  -  & 0.68 $\pm$ 036 \\
%$200$    &  -  & 0.56 $\pm$  0.036 \\
%$500$    &  -  & 0.22 $\pm$ 0.03 \\
%\bottomrule}
%\end{tabular}
%\end{sc}
%\end{small}
%\label{fidelity-table}
%\end{center}
%\vskip -0.1in
%\end{table}

\section{Conclusion}
\label{conclusion}
In this study, we have presented experimental results showing that instances generated by LIME's perturbation procedure are significantly different from training instances drawn from the underlying distribution. Our method of choice for detecting the shift is the MMD kernel two sample kernel test.  The experimental results also investigated the correlations of the fidelity of interpretable model obtained by LIME with the shift. In some cases, the shift can have a negative correlation on the fidelity of the interpretable model. Based on the results from the tests, we argue that random perturbation of features of the explained instance cannot be considered a reliable method of data generation in the LIME framework.

In order to mitigate these shifts, alternative approaches to the perturbation method in LIME may be considered. One natural alternative would be to consider sampling from training instances in the vicinity of the explained instance rather than perturbing the features of the instance.

\subsubsection*{Acknowledgment}
Amir Hossein Akhavan Rahnama would like to thank Patrik Tran and Amir Payberah for their valuable feedback in the process of writing this paper. 

\bibliography{neurips_2019} 
\small

\section*{A. Interpretable Machine Learning}
Many state-of-the-art machine learning models are essentially black-boxes, in the sense that they have a highly nonlinear internal structure that makes a global understanding of their predictions hardly possible in practice. Post-hoc interpretable methods are used to explain the predictions of such black-box models after they are trained. A subclass of post-hoc interpretable methods is called surrogate methods, by which the black-box model is approximated with another (interpretable) model to provide a better understanding of the former. A recent example of these types of method is the popular LIME approach, which is the focus of this work.

\section*{B. Shift in Machine Learning}
A standard assumption in machine learning is that training and test data are coming from the same distribution, namely $P_\text{train}(x, y) \overset{D}{=} P_\text{test}(x, y)$, where $P_\text{train}(x,y)$ and $P_\text{test}(x, y)$ denote the joint probability of an instance $x$ and a label $y$. Since $P(x, y) = P(y|x) P(x)$, any inequality between the two joint distributions may come either from a shift of the prior (covariate shift), i.e., $P_\text{train}(x) \overset{D}{\neq} P_\text{test}(x)$, the conditional, i.e., $P_\text{train}(y|x) \overset{D}{\neq} P_\text{test}(y|x)$ or both (concept drift). \citep{quionero2009dataset}.

\section*{C. Maximum Mean Discrepancy}
Assume that $X$ and $Y$ are random variables defined on a topological space $\chi$, with corresponding Borel probability measures $p$ and $q$. Let us assume we have the independently and identically distributed observations $X := \{x_1, \dots, x_m\}$ and $Y := \{y_1, \dots, y_n\}$\footnote{This notation should not be confused with common usage of $X$ and $Y$ in supervised machine learning literature}. 

Let $\mathcal{F} \colon \chi \rightarrow \mathcal{R}$ be a class of functions. MMD is defined as follows \citep{gretton2012kernel}:

\begin{equation}
\text{MMD}_b[\mathcal{F}, X, Y] := \underset{f \in \mathcal{F}}{\text{sup}}(\mathrm{E}_x[f(x)] -\mathrm{E}_y[f(y)])
\end{equation}

We extensively use the following two properties of MMD that are outlined in Theorem 1 and 2. For proof and a detailed description of the underlying assumptions, see \citep{gretton2012kernel}.

\begin{theorem}
Let $\mathcal{F}$ is a unit ball in a universal Reproducing kernel Hilbert space (RKHS) and is defined on a compact metric space and has a corresponding continuous kernel k(., .). Then $\text{MMD}[\mathcal{F}, p, q] = 0$ if and only if $p = q$.
\end{theorem}

\begin{theorem}
A hypothesis test of level $\alpha$ for the null hypothesis $p=q$, that is, for $\text{MMD}[\mathcal{F},p,q] =0$, has the acceptance region $\text{MMD}_b[F,X,Y]<\sqrt{2K/m}(1+\sqrt{2 \log \alpha^-1})$.
\end{theorem}

Theorem 1 ensures that when MMD is zero, two distributions are equal. Theorem 2 provides an acceptable threshold to be used to reject the null hypothesis, i.e. $p = q$. It should be noted that this test requires no further assumption on the type or class of distributions for $p$ and $q$.

\section*{D. LIME}
\label{lime}
The LIME algorithm provides an explanation for an instance $x$; see Algorithm \ref{lime_general_algorithm}). In addition to this instance, other inputs to the algorithm consists of the following: a trained black-box model $f$, a regularization path $\lambda$, the total number of perturbed instances $n$, the number of features in the explanation $k$ and a kernel function $\pi$. LIME starts off by perturbing non-zero features $x$ uniformly at random for $n$ times. These new instances are called $z_i (i=1, \cdots, n)$. The corresponding distances of $z_i$ (stored in the matrix $Z$) from $x$, namely $D$, is calculated using the kernel function, $\pi$. After that, the $Z$ is transformed into a binary representation called $Z'$. In the next step,  $Z'$ is then fed into the regularization path $\lambda$ (stored in the matrix $Z'_\text{reg}$) to reduce the dimensionality of co-linearly dependent features in $Z$. The interpretable model $g$ is then trained using least squares on $Z'_\text{reg}$ as inputs and the output of the black-box on $Z$ with respect to class $y$, namely $f_y(Z)$ as labels. After the interpretable model is fit, the weights are stored as $|W_g|$. At the end, $K$ features of $Z'_\text{reg}$ with the largest absolute weights in $|W_g|$ are returned as explanations. LIME's loss function is used to measure the quality of explanations:

\begin{equation}\label{lime-loss}
\mathcal{L}(f, g, \pi_x) = \sum_{z, z' \in Z} \pi_x(z) (f(z) - g(z'_\text{reg}))^2 
\end{equation}

The loss is minimized when the output of the interpretable model on the interpretable representation, $g(z'_\text{reg})$, has the least difference from the output of the black-box model on $Z$, namely $f(z)$\footnote{We can only speculate that the formula \ref{lime-loss} is a variation of the fidelity of $g$ with regards to $f$, as is not further explained nor evaluated in \citep{ribeiro2016should}}.

\begin{algorithm}[tb]
\caption{Sparse Linear Explanations using LIME}
\label{lime_general_algorithm}
\begin{algorithmic}[1]
    \STATE {\bfseries Input:} Instance being explained $x$
    \STATE {\bfseries Input:} Class label $y$
    \STATE {\bfseries Input:} Black-box model $f$
    \STATE {\bfseries Input:} Regularization path, $\lambda$
    \STATE {\bfseries Input:} Number of samples, $n$
    \STATE {\bfseries Input:} Number of features, $K$
    \STATE {\bfseries Input:} Kernel function, $\pi$
    \STATE {\bfseries Output:} Explanation, $E$
    \STATE {\bfseries Output:} Loss function, $L$
    \FOR{$i \gets 1$ to $n$}
        \STATE $z_i \gets $ Sample a random subset from non-zero features of $x$ and zero out the rest of features
        \STATE $D_i \gets \pi(z_i, x)$
    \ENDFOR
    \STATE $Z' \gets $ Transform $z_1, \ldots, z_n$ into an interpretable representation 
    \STATE $Z'_\text{reg} \gets \lambda(Z')$  \text{apply regularization path on} $Z'$
    \STATE \text{Train} $g$ \text{using least squares with} $g(Z'_\text{reg})$ \text{as inputs and} $f(Z)$ \text{as labels and store the weights as} $W_g$
    \STATE $E = \text{argmax}(|W_g|)_i \text{s.t.}  \thinspace i=1, \cdots, K$
    \STATE $L = D \times (f(z) - g(z'))^2$
\end{algorithmic}
\end{algorithm}

%\begin{itemize}
%    \item \textbf{Hyper-parameters} (line 3-8): The choice of the type of interpretable model, the choice of the regularization path, the efficient total number of perturbed samples, the number of features in the explanation and the choice of the kernel function
%    \item \textbf{Perturbation} (line 11-13): The method by which new instances are generated from the instance being explained, $x_t$
%    \item \textbf{Transformation} (line 15): how the new instances are transformed into an interpretable binary representation, how regularization is applied on the binary transformed sample
%    \item \textbf{Learning} (line 17): The choice of interpretable model and the type of learning that is used in the fitting of the interpretable model
%    \item \textbf{Explanation} (line 18): The way in which explanations are selected and presented to users 
%    \item \textbf{Loss function} (line 19): The quantitative way in which the quality of the explanation are measured
%\end{itemize}

\section*{E. Details of experiments with additional figures and tables}
\subsection*{E1. Text classification with SVM}
\begin{figure}
  \centering
  \includegraphics[width=0.8\linewidth]{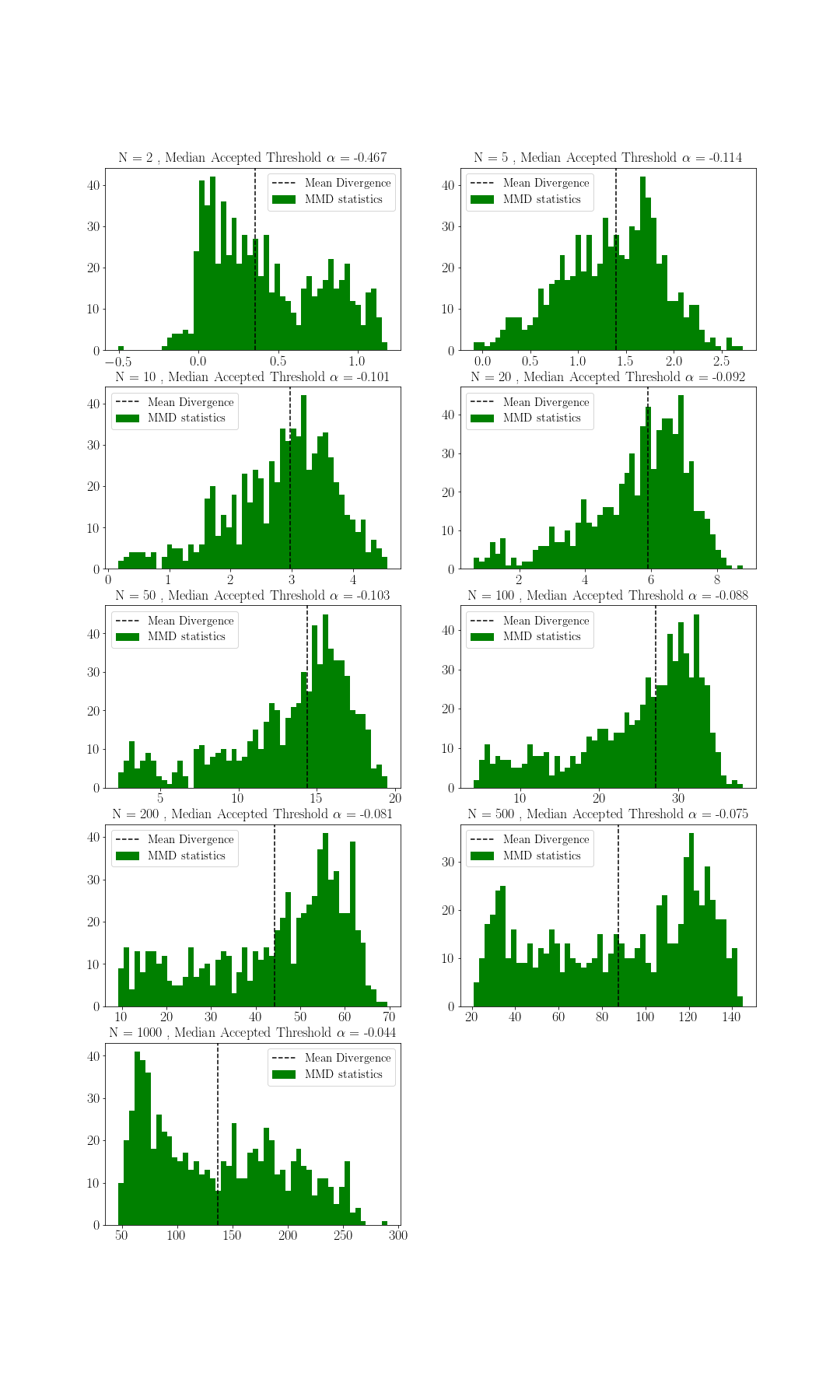}
  \caption{MMD Divergence values for data shift in the Newsgroup dataset}
  \label{data_shift_text_classification}
\end{figure}

\begin{figure}
  \centering
  \includegraphics[width=0.6\linewidth]{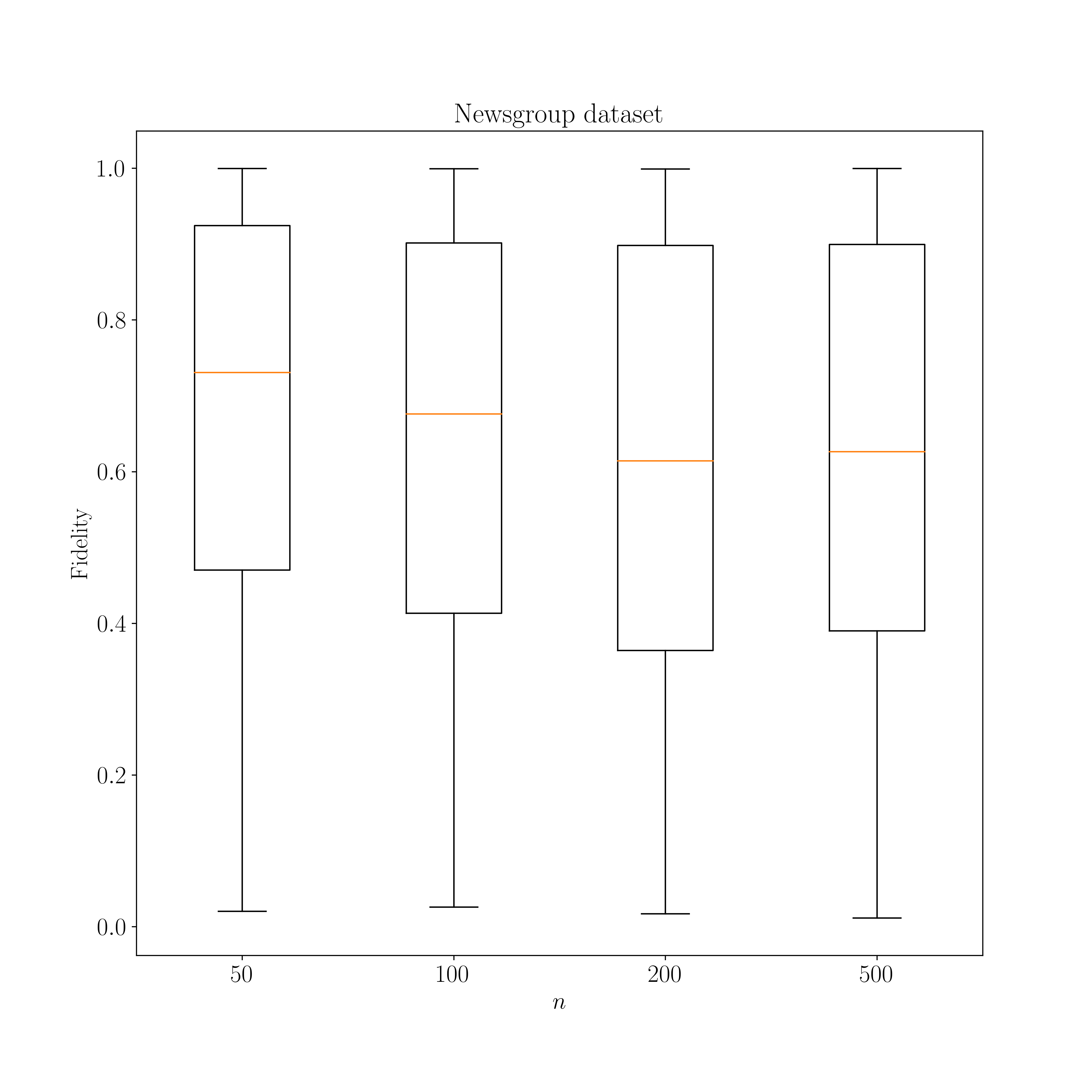}
  \caption{Fidelity of the LIME's interpretable model for explanations of atheism in Newsgroup dataset for each sample size $n$}
  \label{fidelity_text_classification}
\end{figure}

In this experiment, we have replicated the text classification use-case originally investigated in \citep{ribeiro2016should}. The LIME approach is studied together with a black-model consisting of an SVM with a RBF kernel, and using ridge regression as the algorithm for generating surrogate  models. The selected regularization path is Least Angle Regression LASSO, respectively and the kernel function is the cosine kernel. The task concerns binary classification of documents into \textit{christianity} or \textit{atheism}, where the documents come from the newsgroup dataset\footnote{\url{http://qwone.com/~jason/20Newsgroups/}}, divided into 1079 training instances ($X_\text{train}$) and 717 test instances($X_\text{train}$). In this experiment, documents are transformed into the Term-Frequency (TF) representation, corresponding to a total of 19666 words. In this experiment, we limit the explanations to the class label \textit{atheism}. The MMD divergence values for the data shift test, between the perturbed samples of LIME ($Z$) and nearby instances of the explained instance $X_\text{knn}$ can be see in in Figure \ref{data_shift_text_classification} and corresponding divergence values for the label shift divergence measures is shown in Figure \ref{data_shift_text_classification}. In addition, the detailed information on two sample tests between these values for different number of samples ($n$) can be seen in Table \ref{data-shift-table-text-classification}. Lastly, a detailed overview of two sample tests between the predicted value of the black-box model on both samples, namely $f(Z)$ and $f(X_\text{knn})$ is shown in Table \ref{label-shift-table-text-classification}. The boxplot of the fidelity of the interpretable model for various sample size ($n$) can be seen in Figure \ref{fidelity_text_classification} along with mean fidelity values.

\begin{figure}
  \centering
  \includegraphics[width=0.8\linewidth]{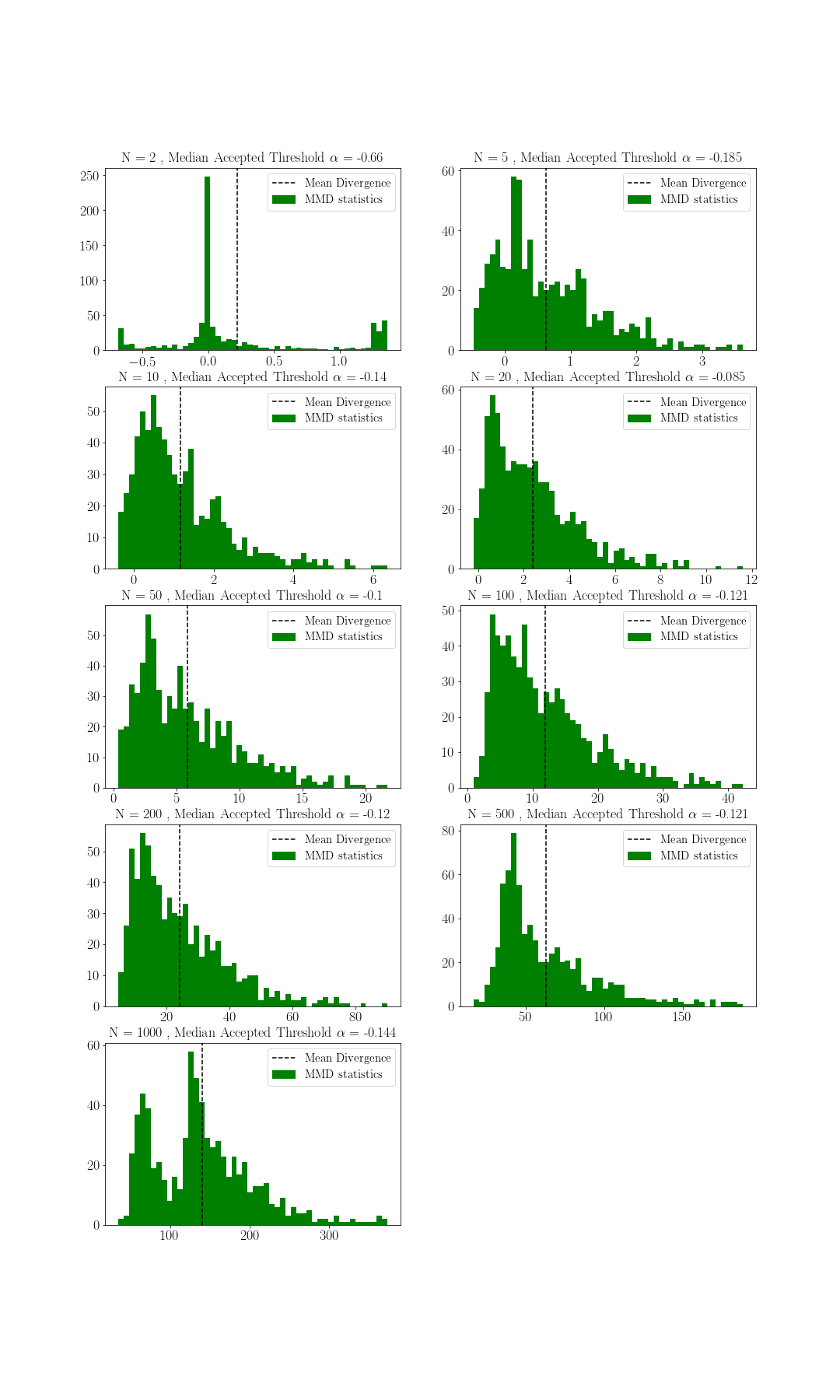}
  \caption{MMD Divergence values for label shift in the Imagenet dataset}
  \label{label_shift_text_classification}
\end{figure}

\begin{table}[t]
\caption{Data shift two sample test results for Newsgroup test instances: $H_0: P_{X_\text{knn}} = P_Z$  ($\alpha=0.05$)}
\label{data-shift-table-text-classification}
\vskip 0.15in
\begin{center}
\begin{small}
\begin{sc}
\begin{tabular}{lcccr}
\toprule
$T_n$ & Reject & Failed to reject & MMD \\
\midrule
$2$    & 417 (57\%)& 300 (43\%) & 0.42 $\pm$ 0.34 \\
$5$    & 681 (94\%)& 36 (6\%) & 1.34 $\pm$ 0.52 \\
$10$    & 710 (99\%)& 7 (1\%) & 2.82 $\pm$ 0.88\\
$20$    & 717 (100\%)& 0 (0\%) & 5.56 $\pm$ 1.58 \\
$50$    & 717 (100\%)& 0 (0\%) & 13.19 $\pm$ 4.03\\
$100$    & 717 (100\%)& 0 (0\%) & 24.77 $\pm$ 8.00 \\
$200$    & 717 (100\%)& 0 (0\%) & 44.20 $\pm$ 15.84 \\
$500$    & 717 (100\%)& 0 (0\%) & 87.35 $\pm$ 36.75 \\
\bottomrule
\end{tabular}
\end{sc}
\end{small}
\end{center}
\vskip -0.1in
\end{table}

\begin{table}[t]
\caption{Label shift two sample test results for Newsgroup test instances along with fidelity measures:$H_0: P_{F(X_\text{knn})} = P_{F(Z)}$ ($\alpha=0.05$)}
\label{label-shift-table-text-classification}
\vskip 0.15in
\begin{center}
\begin{small}
\begin{sc}
\begin{tabular}{lccccr}
\toprule
$T_n$ & Reject & Failed to reject & MMD\\
\midrule
$2$    & 239 (33.3\%)& 478 (66.6\%) & 0.21 $\pm$ 0.56 \\
$5$    & 209 (29.1\%)& 508 (70\%) & 0.62 $\pm$ 0.78 \\
$10$    & 336 (46.8)& 381 (1\%) & 1.16 $\pm$ 1.15 \\
$20$    & 515 (71.8\%)& 202 (28.1\%) & 2.38 $\pm$ 1.93  \\
$50$    & 697 (85.2\%)& 20 (0.02\%) & 5.88 $\pm$ 4.00  \\
$100$    & 716 (99.8\%)& 1 (0.2\%) & 11.97 $\pm$ 7.69  \\
$200$    & 717 (100\%)& 0 (0\%) & 24.29 $\pm$ 14.47 \\
$500$    & 717 (100\%)& 0 (0\%) & 63.06 $\pm$ 31.16  \\
\bottomrule
\end{tabular}
\end{sc}
\end{small}
\label{concept_drift_text_classification_table}
\end{center}
\vskip -0.1in
\end{table}

\subsection*{E2. Object detecton with Deep Neural Networks}
\begin{figure}
  \centering
  \includegraphics[width=0.6\linewidth]{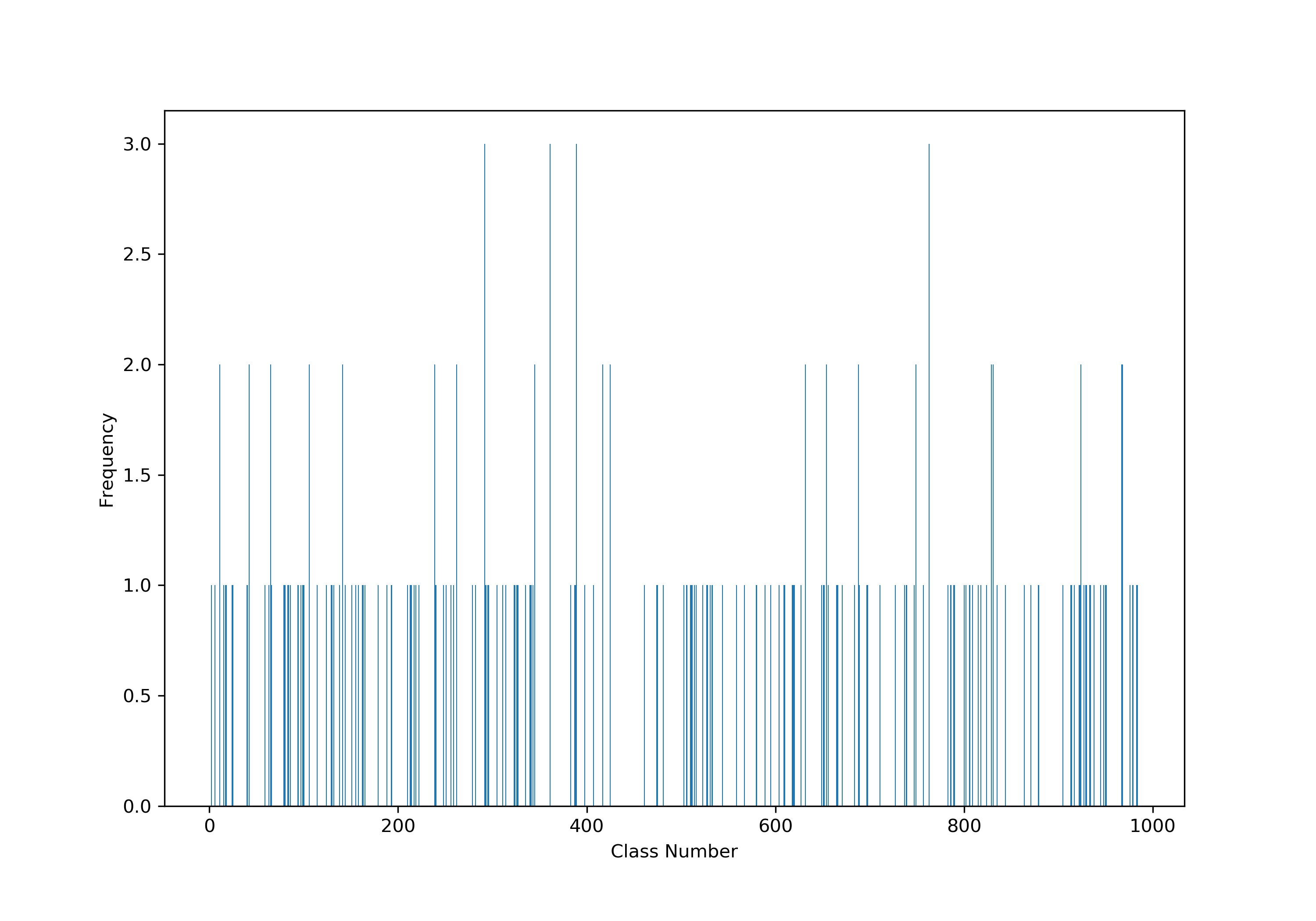}   
  \caption{Predicted Class Frequency of our ImageNet sample by Inception V3}
  \label{predicted_class_frequency}
\end{figure}

In this experiment, we have replicated the objected detection use-case originally investigated in LIME's paper. The black-model is the pre-trained Inception V3 model, and ridge regression as the model for the surrogate. The selected regularization path is Least Angle Regression LASSO, and the kernel function is the cosine kernel.  The dataset for this experiment is ImageNet dataset. The explained image is transformed into the super-pixels representation using Quickshift algorithm. The task concerns the multi-class object detection of images in the Imagenet dataset with $1001$ classes. The dataset has $10000000$ images in its training set, namely ($X_\text{train}$) and $60000$ in its test set, namely ($X_\text{train}$). In our experiment, LIME explanations for the top predicted class are considered. In this use-case we have selected a sample of 200 instances from Imagenet (see Figure \ref{predicted_class_frequency} for predicted class distribution of the instances).

\begin{figure}
  \centering
  \includegraphics[width=0.8\linewidth]{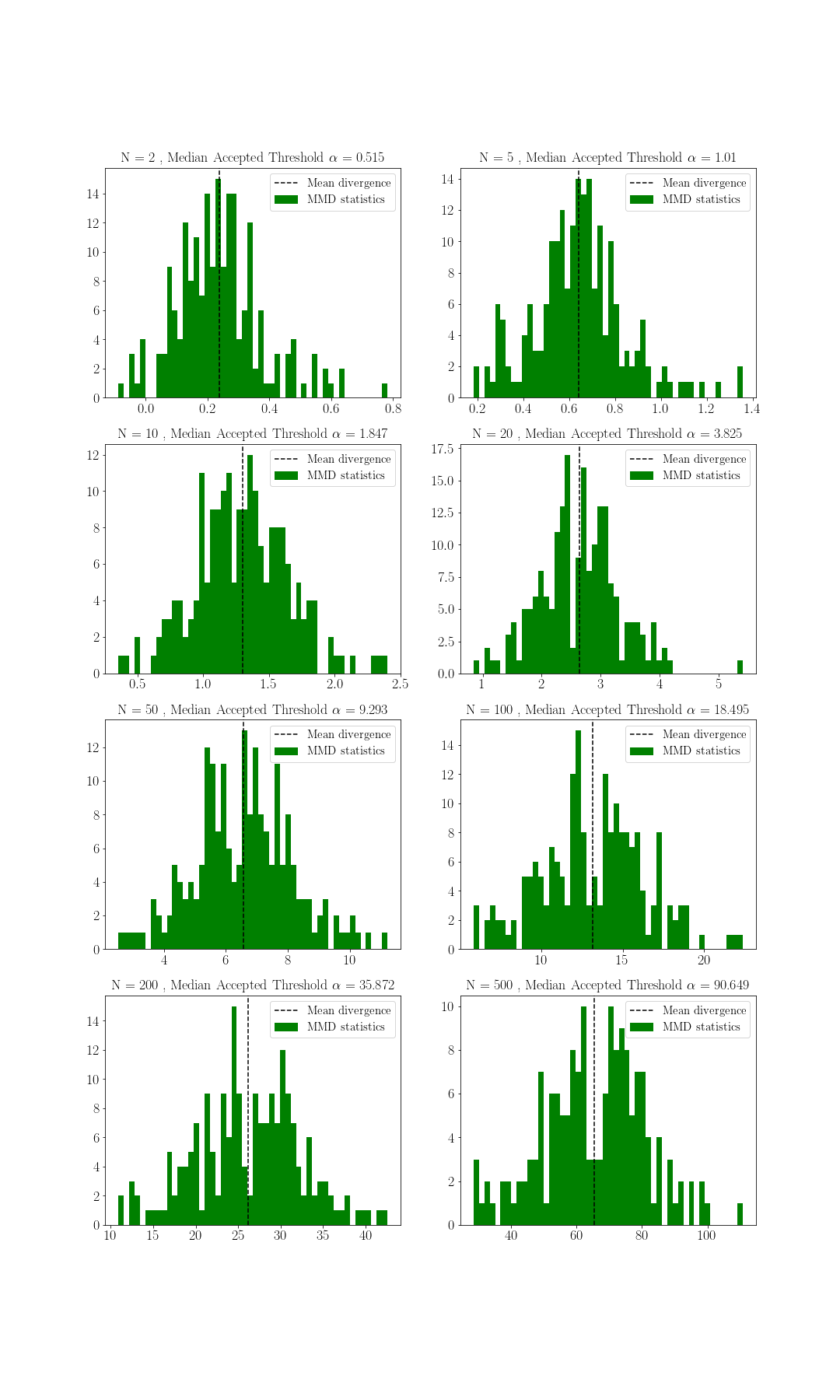}
  \caption{MMD Divergence values for data shift in the ImageNet dataset}
  \label{data_shift_object_detection}
\end{figure}

\begin{figure}
  \centering
  \includegraphics[width=0.8\linewidth]{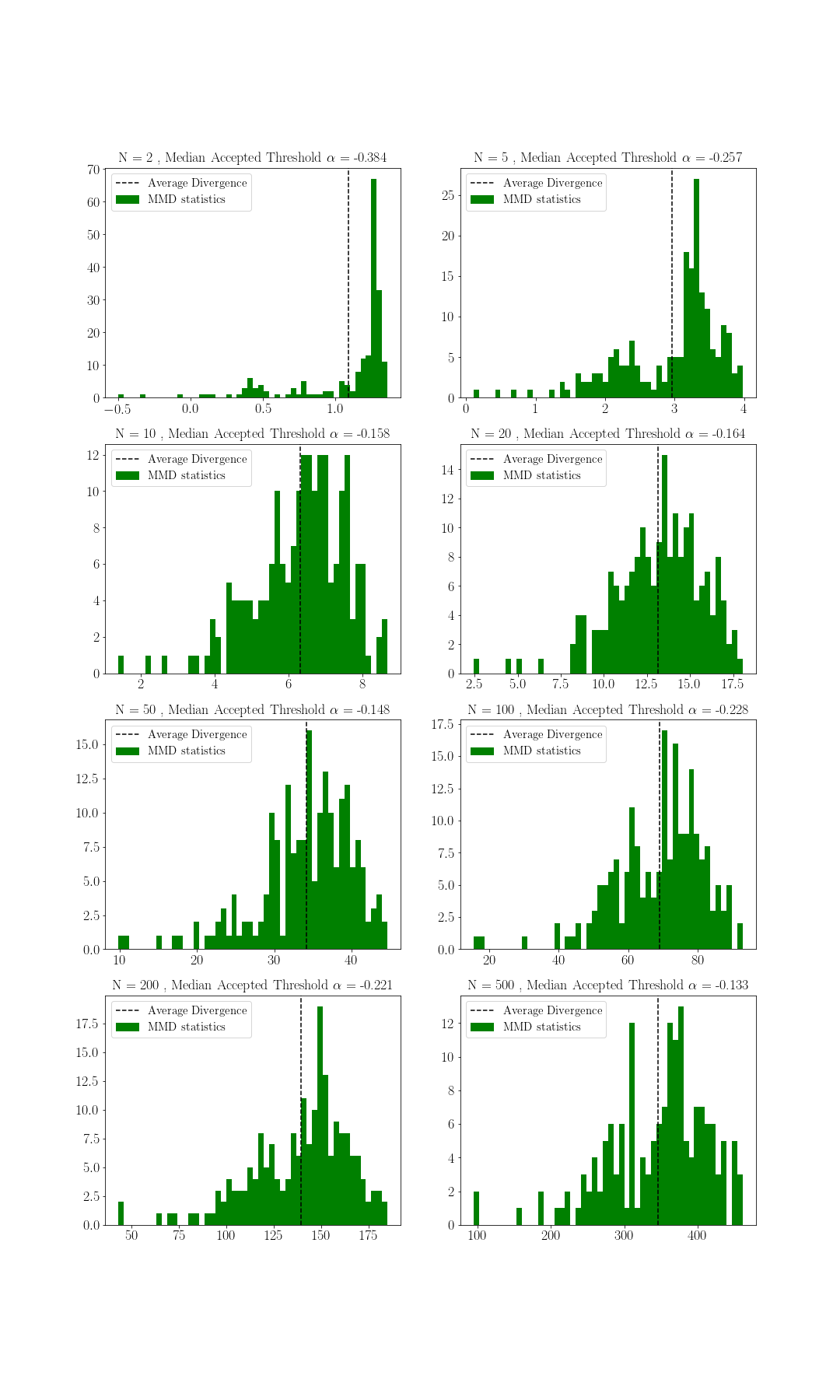}
  \caption{MMD Divergence values for label shift in the ImageNet dataset}
  \label{label_shift_object_detection}
\end{figure}

\begin{figure}
  \centering
  \includegraphics[width=0.6\linewidth]{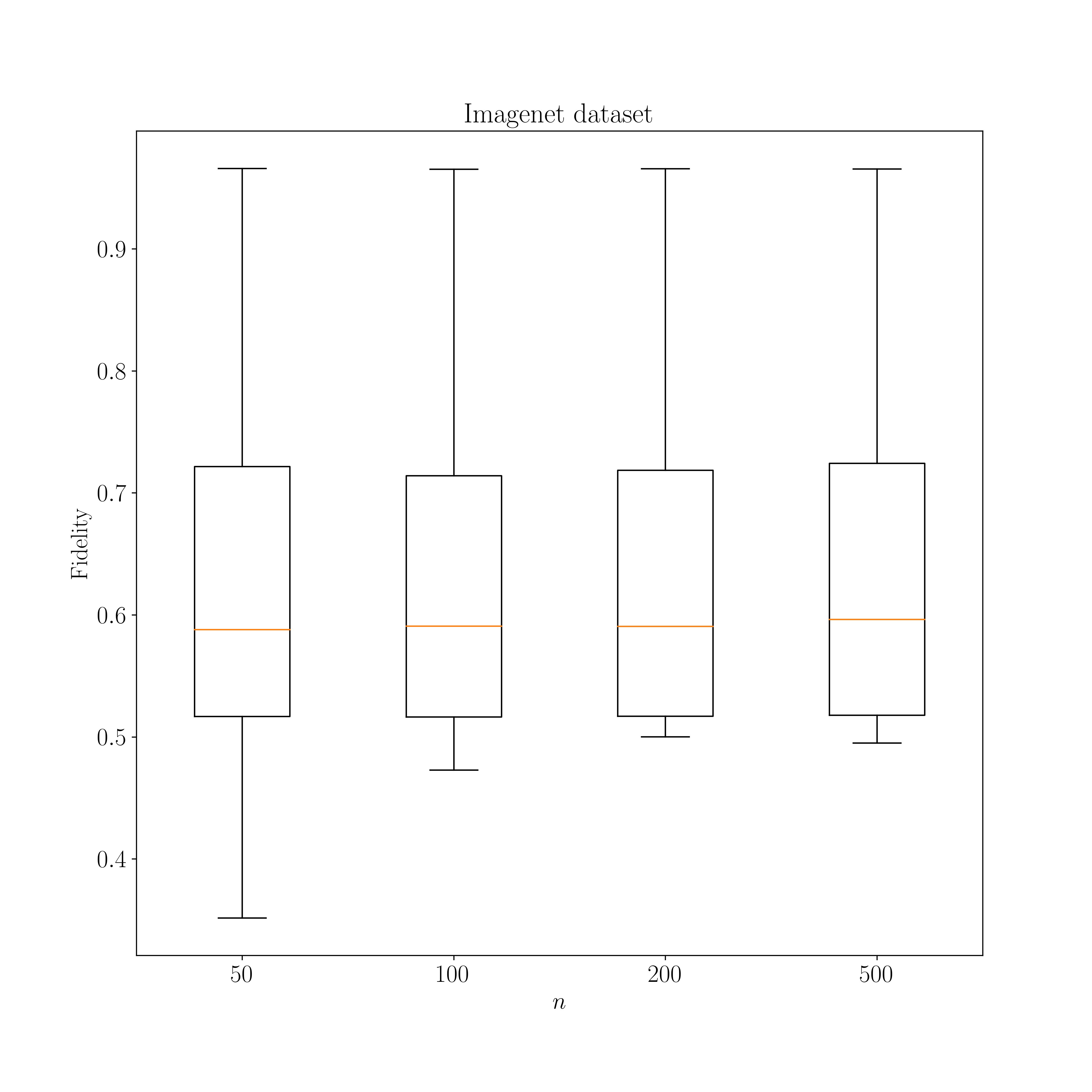}   
  \caption{Fidelity histogram for LIME explanations of top-1 predicted class in ImageNet}
  \label{fidelity_object_detection}
\end{figure}

\begin{figure}
  \centering
  \includegraphics[width=0.8\linewidth]{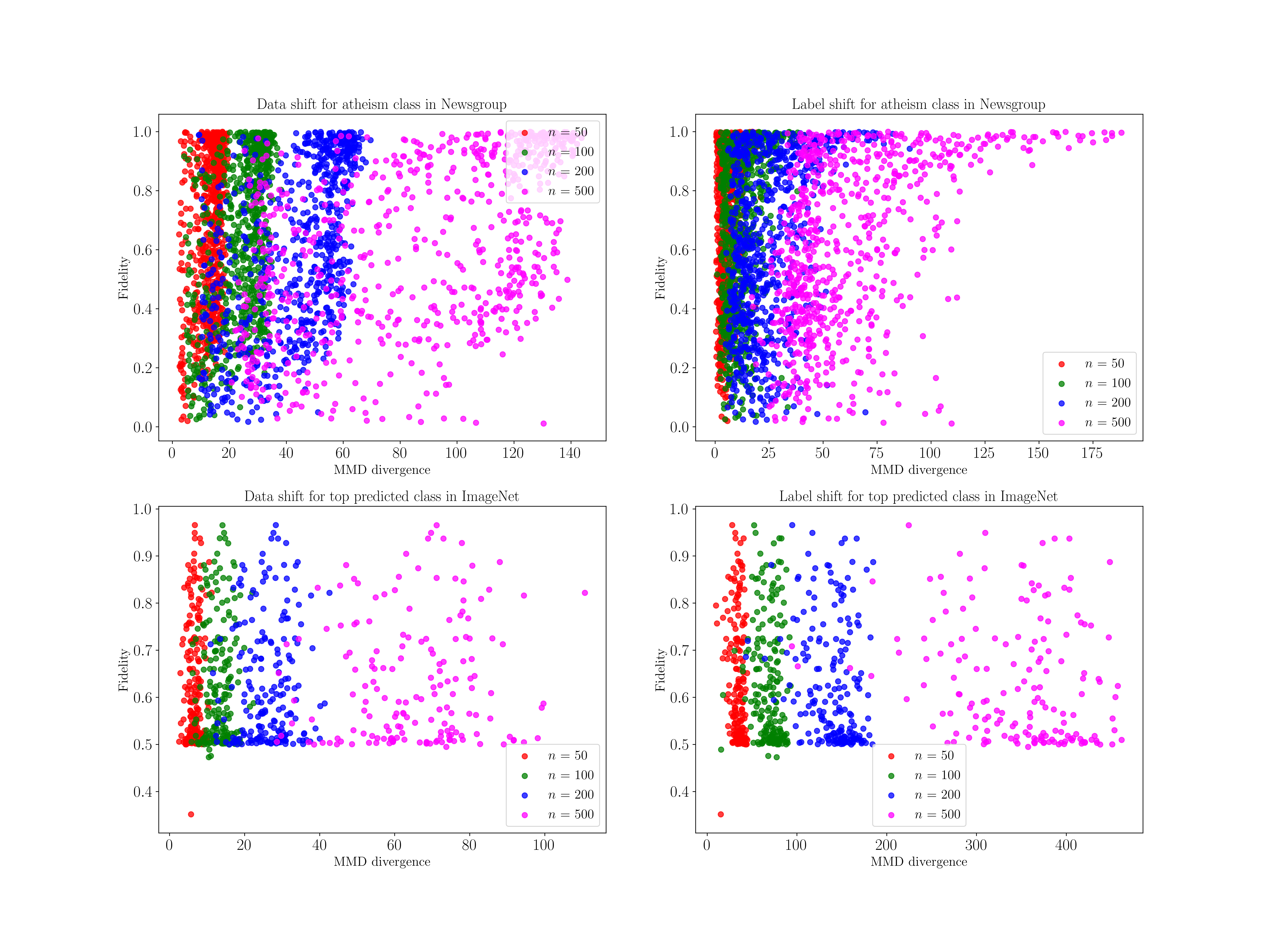} 
  \caption{MMD divergence and fidelity in both use-cases}
  \label{fidelity_object_detection_text_classification}
\end{figure}

The MMD divergence values for the data shift test, between the perturbed samples of LIME ($Z$) and nearby instances of the explained instance $X_\text{local}$ can be see in in Figure \ref{data_shift_object_detection} and corresponding divergence values for the label shift divergence measures is shown in Figure \ref{label_shift_object_detection}. The boxplot of the fidelity of the interpretable model for various sample size ($n$) can be seen in Figure \ref{fidelity_object_detection} along with mean fidelity values. In Figure \ref{fidelity_object_detection_text_classification}, the relationship between MMD divergence values and fidelity in both tests are shown.

\begin{table}[t]
\caption{Data shift two sample test results for ImageNet test instances for their top-1 predicted class label: $H_0: P_{X_\text{local}} = P_{Z}$ ($\alpha=0.05$)}
\vskip 0.15in
\begin{center}
\begin{small}
\begin{sc}
\begin{tabular}{lcccr}
\toprule
$n$ & Reject & Failed to reject & MMD \\
\midrule
$2$    & 86 (43 \%)& 113 (56\%) & 0.23 $\pm$ 0.13 \\
$5$    & 188 (100 \%)& 0 (0\%) & 0.64 $\pm$ 0.20 \\
$10$    & 188 (100 \%)& 0 (0\%) & 1.29 $\pm$ 0.35 \\
$20$    & 188 (100 \%)& 0 (0\%) & 2.63 $\pm$ 0.67 \\
$50$    & 188 (100 \%)& 0 (0\%) & 6.56 $\pm$ 0.13 \\
$100$    & 188 (100\%)& 0 (0\%) & 13.16 $\pm$ 0.20 \\
$200$    & 188 (100\%)& 0 (0\%) & 26.21 $\pm$ 0.35 \\
$500$    & 188 (100\%)& 0 (0\%) & 65.32 $\pm$ 0.67 \\
\bottomrule
\end{tabular}
\end{sc}
\end{small}
\label{data_shift_object_detection_table}
\end{center}
\vskip -0.1in
\end{table}

\begin{table}[t]
\caption{Label shift two sample test results for ImageNet test instances for their top-1 predicted class label: $H_0: P_{F(X_{\text{local}})} = P_{F(Z)}$ ($\alpha=0.05$)}
\vskip 0.15in
\begin{center}
\begin{small}
\begin{sc}
\begin{tabular}{lcccr}
\toprule
$n$ & Reject & Failed to reject & MMD \\
\midrule
$2$    & 188 (100 \%)& 0 (0\%) & 1.08 $\pm$  0.34 \\
$5$    & 188 (100 \%)& 0 (0\%) & 2.96 $\pm$ 0.71 \\
$10$    & 188 (100 \%)& 0 (0\%) & 6.31 $\pm$ 1.23 \\
$20$    & 188 (100 \%)& 0 (0\%) & 13.16 $\pm$ 2.58 \\
$50$    & 188 (100 \%)& 0 (0\%) & 34.18 $\pm$ 6.13 \\
$100$    & 188 (100\%)& 0 (0\%) & 69.03 $\pm$ 12.73 \\
$200$    & 188 (100\%)& 0 (0\%) & 139.16 $\pm$ 25.63 \\
$500$    & 188 (100\%)& 0 (0\%) & 346.30 $\pm$ 67.99 \\
\bottomrule
\end{tabular}
\end{sc}
\end{small}
\label{label_shift_object_detection_table}
\end{center}
\vskip -0.1in
\end{table}

\end{document}